\documentclass[conference]{IEEEtran}
\IEEEoverridecommandlockouts
\usepackage{cite}
\usepackage{amsmath,amssymb,amsfonts}
\usepackage{algorithmic}
\usepackage[ruled,linesnumbered,noend]{algorithm2e} 
\usepackage{subcaption} 
\usepackage{booktabs} 
\usepackage{hyperref}
\usepackage{graphicx}
\usepackage{textcomp}
\usepackage{amssymb}
\usepackage{multirow}
\usepackage{amsmath}
\usepackage{float}
\usepackage{bm}
\usepackage{xcolor}
\usepackage{enumitem}
\usepackage{fancyhdr}

\DeclareMathOperator*{\argmin}{arg\,min}
\def\BibTeX{{\rm B\kern-.05em{\sc i\kern-.025em b}\kern-.08em
    T\kern-.1667em\lower.7ex\hbox{E}\kern-.125emX}}
\begin{document}

\pagestyle{fancy}
\fancyhead[L]{Accepted by IEEE ICDM 2022 as a conference paper} 
\newcommand{\jy}[1]{\textcolor{black}{#1}}
\newcommand{\dk}[1]{\textcolor{black}{#1}}
\newcommand{\qj}[1]{\textcolor{black}{#1}}

\title{Conflict-Aware Pseudo Labeling via Optimal Transport for Entity Alignment\\}

\author{\IEEEauthorblockN{Qijie Ding}
\IEEEauthorblockA{\textit{Discipline of Business Analytics} \\
\textit{The University of Sydney}\\
Sydney, Australia \\
qijie.ding@sydney.edu.au}
\and
\IEEEauthorblockN{Daokun Zhang}
\IEEEauthorblockA{\textit{Department of Data Science \& AI} \\
\textit{Monash University}\\
Melbourne, Australia \\
daokun.zhang@monash.edu}
\and
\IEEEauthorblockN{Jie Yin}
\IEEEauthorblockA{\textit{Discipline of Business Analytics} \\
\textit{The University of Sydney}\\
Sydney, Australia \\
jie.yin@sydney.edu.au}
}

\maketitle

\begin{abstract}
Entity alignment aims to discover unique equivalent entity pairs with the same meaning across different knowledge graphs (KGs). 
Existing models have focused on projecting KGs into a latent embedding space so that inherent semantics between entities can be captured for entity alignment. However, the adverse impacts of alignment conflicts have been largely overlooked during training, thereby limiting the entity alignment performance. 
To address this issue, we propose a novel \underline{C}onflict-aware \underline{P}seudo \underline{L}abeling \qj{via \underline{O}ptimal \underline{T}ransport} model (CPL-OT) for entity alignment. The key idea is to iteratively pseudo-label alignment pairs empowered with conflict-aware optimal transport (OT) modeling to boost the precision of entity alignment. CPL-OT is composed of two key components---entity embedding learning with global-local aggregation and iterative conflict-aware pseudo labeling---that mutually reinforce each other. To mitigate alignment conflicts during pseudo labeling, we propose to use optimal transport as an effective means to warrant one-to-one entity alignment between two KGs with the minimal overall transport cost. Extensive experiments on benchmark datasets validate the superiority of CPL-OT over state-of-the-art baselines under both settings with and without prior alignment seeds. 

\end{abstract}

\begin{IEEEkeywords}
knowledge graph, entity alignment, pseudo labeling, optimal transport
\end{IEEEkeywords}

\section{Introduction}
Knowledge Graphs (KGs) comprise of graph-structured semantic information about real-world concepts (or entities) and relations among these concepts. KGs are widely adopted in various AI-powered applications to provide strong inference capabilities. 
Yet, it is well recognized that real-world KGs suffer from incompleteness arising from their complex, semi-automatic construction processes. 
To enrich knowledge representation over incomplete KGs, entity alignment aims to link entities with the same real-world identity across KGs. 


Mainstream entity alignment models are based on KG embedding, which embeds KGs into a latent vector space to capture inherent semantics regardless of the heterogeneity among KGs. To learn better KG embeddings, methods like GCN-Align~\cite{wang2018cross} leverage graph convolutional networks (GCNs)~\cite{kipf2016semi} to capture structural and neighboring entity information for entity alignment. Recent studies~\cite{wu2019jointly,wu2019relation,zhu2021relation2} utilize a highway strategy~\cite{srivastava2015highway} to alleviate the over-smoothing issue during GCN propagation, or jointly learn entity and relation embeddings for improving the precision of entity alignment. Other works tackle the shortage of pre-aligned entity pairs (known as \textit{prior alignment seeds}) provided for training. 
BootEA~\cite{sun2018bootstrapping}, IPTransE~\cite{zhu2017iterative} and MRAEA~\cite{mao2020mraea} propose bootstrapping strategies to iteratively augment alignment seeds for improving subsequent training. 

Despite making remarkable progress, current methods have largely overlooked the adverse impacts of alignment conflicts during training, i.e., multiple entities in one KG are simultaneously aligned with a single entity in another KG. The presence of alignment conflicts is mainly due to two reasons. First, although graph convolution enables to effectively encode entity neighborhood information into entity embeddings, GCN-based methods often incur more conflicting alignment pairs due to the innate feature smoothing effect. Second, 
conflicting alignment pairs would adversely impair the quantity of correctly pseudo-labeled alignment pairs, thus jeopardizing the efficacy of subsequent model training. This restricts the performance of pseudo labeling based KG alignment.

In this paper, we propose a novel \underline{C}onflict-aware \underline{P}seudo \underline{L}abeling \qj{via \underline{O}ptimal \underline{T}ransport} model (CPL-OT) for entity alignment. 
Our core idea is to pseudo-label alignment pairs via conflict-aware OT modeling to boost the precision of entity alignment. CPL-OT consists of two key components---entity embedding learning with global-local aggregation and iterative conflict-aware pseudo labeling---that alternately reinforce each other. Specifically, we make the following contributions.

\begin{itemize}[leftmargin=*]

\item We propose an iterative conflict-aware pseudo labeling strategy that selects the most reliable alignment pairs via OT modeling. 
The OT models entity alignment as a process of transporting each entity in one KG to a unique entity in another KG with the minimal overall transport cost, warranting one-to-one entity alignment. 

\item We design graph convolution with global-local aggregation for learning expressive entity embeddings. 
The rectified distance between entity embeddings are used as the transport cost for OT modeling to mitigate alignment conflicts during pseudo labeling. 

\item Experimental results on benchmark datasets show that CPL-OT yields competitive results with or without prior alignment seeds, outperforming state-of-the-art baselines. 

\end{itemize}



\section{Related Work}
We review two streams of related work: entity alignment in knowledge graphs and optimal transport on graphs.

\subsection{Entity Alignment in Knowledge Graphs}
Most entity alignment models are embedding-based approaches, which embed KGs into a unified vector space by imposing the embeddings of pre-aligned entity pairs to be as close as possible. This ensures that alignment similarities between entities can be directly measured via their embeddings.


To leverage KG structural information, methods like GCN-Align~\cite{wang2018cross} utilize GCNs to learn better entity embeddings for alignment. However, GCNs and their variants \dk{are inclined to result in alignment conflicts, because their feature smoothness schemes make entities have similar embeddings among local neighborhoods. To alleviate the over-smoothing issue,} 
recent works~\cite{wu2019jointly,wu2019relation,zhu2021relation2} adopt a highway strategy~\cite{srivastava2015highway} on GCN layers, \jy{which ``mixes" the smoothed entity embeddings with the original features.}
Other models such as HGCN~\cite{wu2019jointly}, RDGCN~\cite{wu2019relation}, and RNM~\cite{zhu2021relation2} consider relations in KGs to reinforce GCN-based entity embeddings.
Nonetheless, these models require an abundance of prior alignment seeds for training, which are labor-intensive and costly to obtain in real-world KGs. 

To tackle the shortage of prior alignment seeds, semi-supervised methods such as BootEA~\cite{sun2018bootstrapping}, IPTransE~\cite{zhu2017iterative}, RNM~\cite{zhu2021relation2}, \qj{and MRAEA~\cite{mao2020mraea}} propose bootstrapping strategies to iteratively augment alignment seeds. These models, however, inevitably introduce alignment conflicts during bootstrapping, as they sample possible alignment pairs directly based on embedding distances. To handle alignment conflicts, RNM~\cite{zhu2021relation2} and MRAEA~\cite{mao2020mraea} use simple heuristics to preserve only the most convincing alignment pairs. 
BootEA~\cite{sun2018bootstrapping} adopts a bipartite graph max-weighted matching strategy to select a small number of the most likely aligned pairs at each iteration, and then accumulates pseudo labels across  iterations, which inevitably incurs alignment conflicts. 
In our work, we model entity alignment as an OT process, warranting 
a larger quantity of correctly aligned entity pairs to be pseudo-labeled at each iteration without conflicts. This offers sufficient supervision to learn informative entity embeddings for alignment inference.


\subsection{\dk{Optimal Transport on Graphs}}
Optimal transport (OT) aims to find an optimal plan to move one distribution of mass to another with the minimal cost~\cite{villani2009optimal}. 
Recently, OT has been studied for cross-lingual KG entity alignment~\cite{pei2019improving} and cross-domain alignment on graphs~\cite{chen2020graph}. The transport on the edges across graphs has also been used to define the Gromov-Wasserstein distance to measure graph matching similarity~\cite{xu2019gromov} or to boost the entity alignment performance~\cite{chen2020graph}. However, these methods have primarily used OT to define a learning objective, involving bi-level optimization for model training with high computational cost. Thus, they cannot be directly applied to our context of iterative pseudo labeling. 

\section{Problem Definition}
A knowledge graph can be denoted as $G = \{E, R, T\}$ with the entity set $E$, relation set $R$ and triplet set $T$. We use $e \in E$, $r \in R$, $(e_{i}, r, e_{j}) \in T$ to represent an entity, a relation and a triplet, respectively. Each entity $e$ is characterized by a feature vector $\bm{x}_e \in \mathbb{R}^{d}$, which can be obtained from its textual descriptions or entity name with semantic meanings. Formally, two individual KGs are given for the task of entity alignment, i.e., $ G_{1} = \{E_{1}, R_{1}, T_{1}\}$ and $G_{2} = \{E_{2}, R_{2}, T_{2}\}$.
An entity $e_i\in E_{1}$ in $G_1$ is likely to correspond to the same concept with another entity $e_j\in E_2$ in $G_2$, denoted as $e_{i} \Leftrightarrow e_{j}$, and vice versa.

To provide supervision for entity alignment, a small number of pre-aligned entity pairs between $G_{1}$ and $G_{2}$ are sometimes provided as prior alignment seeds in the form of $\mathbb{L}^0_e=\{ (e_{i}, e_{j}) | e_{i} \in E_1, e_{j} \in E_2, e_{i} \Leftrightarrow e_{j}\}$. In some cases, prior alignment seeds may be unavailable due to high labeling cost, such that $\mathbb{L}^0_e=\emptyset$. Along with prior alignment seeds, there are two sets of unaligned entities $E'_1\subseteq E_1$ and $E'_2\subseteq E_2$ in two KGs, with $E'_1= E_1$ and $E'_2= E_2$ when $\mathbb{L}^0_e=\emptyset$. The task of entity alignment is to discover unique equivalent entity pairs $(e_i,e_j)$ with $e_i\in E'_1$, $e_j\in E'_2$ and $e_{i} \Leftrightarrow e_{j}$ across $G_1$ and $G_2$, based on prior alignment seeds $\mathbb{L}^0_e$, KG structure, and entity features in $G_1$ and $G_2$.

\section{The Proposed Method}

\dk{To effectively perform entity alignment with the shortage of prior alignment seeds, the proposed CPL-OT model uses an OT-based pseudo labeling to augment entity alignment seeds and provide more supervisions for entity alignment inference. 
CPL-OT comprises of two components: global-local aggregation for entity embedding and conflict-aware pseudo labeling for alignment augmentation. } The two components are performed alternately in an iterative way until convergence.



\subsection{Global-Local Aggregation for Entity Embedding}
To leverage \dk{relational structures}, we conduct two levels of neighborhood aggregation for each entity, i.e., global-level relation aggregation and local-level entity aggregation. 

\subsubsection{Global-Level Relation Aggregation} First, for each relation $r_i\in R_1 \cup R_2$, we construct its feature \dk{vector} $\bm{x}_{r_i}$ as the \dk{averaged} concatenation of the feature \dk{vectors} of its associated head and tail entities:
\begin{equation}
\bm{x}_{r_i} = \frac{1}{|\{(e_h, r_i, e_t) \in T_1 \cup T_2\}|} \sum_{(e_h, r_i, e_t) \in T_1 \cup T_2} [\bm{x}_{e_h}\| \bm{x}_{e_t}],
\end{equation}
where $[\cdot\|\cdot]$ denotes the concatenation operation, \begin{math} \{(e_h, r_i, e_t) \in T_1 \cup T_2\} \end{math} is the set of all triplets containing relation \begin{math} r_i \end{math}, $\bm{x}_{e_h}$ and $\bm{x}_{e_t}\in\mathbb{R}^{d}$ are the feature vectors of entity $e_h$ and $e_t$, respectively. 
Then, for each entity $e_i\in E_1 \cup E_2$, we construct its \dk{averaged} neighboring relation feature vector as
\begin{equation}
  \bm{x}_{e_i\_rels} = \frac{1}{|\mathcal{N}_r(e_i)|} \sum_{r_j \in \mathcal{N}_r(e_i)} \mathbb{I}_{e_i}(r_j) \cdot \bm{x}_{r_j},
\end{equation}
where $\mathcal{N}_r(e_i)$ is the set of one-hop neighboring relations of entity $e_i$, and $\mathbb{I}_{e_i}(r_j)$ indicates \dk{the direction of relation $r_j$ with regards to $e_i$}, with \dk{$-1$ for $e_i$ being the successor} and $+1$ for \dk{$e_i$ being the predecessor}. \dk{The consideration of the direction can incorporate richer relational neighborhood structures.} 

To perform global-level relation aggregation, we concatenate each entity's averaged neighboring relation feature vector $\bm{x}_{e_i\_rels}\in \mathbb{R}^{2d}$ with its original feature vector $\bm{x}_{e_i} \in \mathbb{R}^d$, \dk{followed by} a non-linear transformation:
\begin{equation}
  \bm{h}
  _{e_i}^{(1)} = \mathrm{ReLU}(W_1[\bm{x}_{e_i} \| \bm{x}_{e_i\_rels}]+b_1) + \bm{x}_{e_i},
\end{equation}where 
$W_1\in\mathbb{R}^{d\times 3d}$ and $b_1\in\mathbb{R}^d$ are the \dk{weight} matrix and the bias \dk{vector}, respectively.
To avoid over-smoothing, 
we add back \dk{
the original entity feature vector $\bm{x}_{e_i}$.}

\subsubsection{Local-Level Entity Aggregation} After obtaining relation aggregated entity embeddings, we conduct local-level entity aggregation to capture neighboring entity structure.

To this end, we take advantage of a two-layer GCN~\cite{kipf2016semi} together with a highway gate strategy~\cite{srivastava2015highway} to 
\dk{avoid over-smoothing.}
Formally, we first stack relation aggregated entity embeddings $\bm{h}_{e_i}^{(1)}$ for each entity $e_i\in E_1\cup E_2$ into an embedding matrix $H^{(1)}_e\in\mathbb{R}^{(|E_1|+|E_2|)\times d}$. 
Then, the entity embedding matrix $H^{(1)}_e$ is updated as follows \dk{from} $l=1$:
\begin{equation}
\left\{
\begin{aligned}
&\tilde{H}_{e}^{(l+1)} = \mathrm{ReLU}(\tilde{D}^{-\frac{1}{2}} \tilde{A} \tilde{D}^{-\frac{1}{2}} H_{e}^{(l)} W_{l+1}),\\
&H_{e}^{(l+1)} = T(H_{e}^{(l)}) \odot \tilde{H}^{(l+1)}_{e} + (1-T(H_{e}^{(l)})) \odot H^{(l)}_{e},
\end{aligned}
\right.
\end{equation}
where \begin{math}\tilde{A} = A + I_{|E_1|+|E_2|}\end{math} is the undirected adjacency matrix of the two combined graphs \begin{math}G_1 \cup G_2\end{math} \dk{augmented by} self-connections \dk{that are} represented by the $(|E_1|+|E_2|)\times (|E_1|+|E_2|)$ identity matrix \begin{math} I_{|E_1| +|E_2|}\end{math}, \begin{math}\tilde{D}_{ii} = \sum_{j} \tilde{A}_{ij} \end{math} is the degree matrix, \begin{math} W_{l+1} \in \mathbb{R}^{d\times d}\end{math} is \dk{the weight} matrix \dk{at layer $l$} and \begin{math}\odot\end{math} is the Hadamard product (or element-wise multiplication). Specifically, \begin{math} T(H_{e}^{(l)})\in\mathbb{R}^{(|E_1|+|E_2|)\times d}\end{math} is the transformation gate \dk{obtained from} $H_{e}^{(l)}$. \dk{The use of the transformation gate can effectively filter out \dk{over-smoothed} feature dimensions}. After neighboring entity aggregation, \dk{we can obtain entity embeddings $H_{e}^{(3)}$ for KG alignment}.

Formally, we denote the final embedding for entity $e_i$ as $\bm{h}_{e_i}=\bm{h}^{(3)}_{e_i}\in\mathbb{R}^{d}$, where $\bm{h}^{(3)}_{e_i}\in\mathbb{R}^{d}$ is the transpose of the row vector of $H_{e}^{(3)}$ \dk{indexed by} entity $e_i$. 

\subsection{Conflict-Aware Pseudo Labeling for Alignment Augmentation with Optimal Transport}
With the constructed entity embeddings, we perform pseudo labeling to identify new reliable alignment pairs. A simple approach is to calculate the embedding distance $d(e_i,e_j)$ between each entity pair $(e_i,e_j)$ across two KGs: 
\begin{equation}
  d(e_i,e_j)=\|\bm{h}_{e_i}-\bm{h}_{e_j}\|_1,
\label{eq:dist}
\end{equation}\dk{where $\|\cdot\|$ denotes 1-norm.}
The entity pairs with a distance smaller than a pre-defined threshold are then identified as the pseudo-labeled alignment pairs. However, this simplistic pseudo labeling approach might lead to \dk{errors}, especially when entity embeddings are not well learned in the presence of scarce prior alignment seeds.

To address this issue, we use the relational neighborhood matching~\cite{zhu2021relation2} 
to rectify embedding based distance in Eq. (\ref{eq:dist}): 
\begin{equation}
\tilde{d}(e_i,e_j)=d(e_i,e_j)-\lambda s(e_i,e_j),
\label{eq:rectify_dist}
\end{equation}where $\lambda$ is a trade-off parameter and $s(e_i,e_j)$ is the relational neighborhood matching similarity~\cite{zhu2021relation2} calculated by comparing neighborhoods, including neighboring entities and neighboring relations, between $(e_i,e_j)$.

For each unaligned entity $e_i\in E'_1$ in $G_1$, we can find its alignment $e_j\in E'_2$ in $G_2$ with the shortest rectified distance: 
\begin{equation}
e_i \Leftrightarrow \argmin_{e_j\in E_2}\tilde{d}(e_i,e_j).
\label{eq:naive_align}
\end{equation}

Due to the smoothing effect of neighborhood aggregation, entities tend to have indistinguishable representations, resulting in alignment conflicts.
To avoid this issue, we propose an OT-based strategy. The aim is to find an optimal plan to transport each unaligned entity in $G_1$ to a unique unaligned entity in $G_2$. As such, a globally optimal alignment configuration can be discovered with the minimal overall inconsistency.

From the unaligned entity sets $E'_1 \subseteq E_1$ and $E'_2 \subseteq E_2$, we first identify the alignment candidates by considering only the cross-KG entity pairs with the rectified distance smaller than a pre-defined threshold $\theta$:
\begin{equation}
\mathcal{C}_{\theta}=\{(e_i,e_j)|e_i\in E'_1, e_j\in E'_2, \; \tilde{d}(e_i,e_j)< \theta\}.
\end{equation}
The entity sets in alignment candidates are then denoted as:
\begin{equation}
\begin{aligned}
&E''_1=\{e_i\in E'_1|\;\exists\; e_j\in E'_2,\;(e_i,e_j)\in \mathcal{C}_{\theta}\},\\
&E''_2=\{e_j\in E'_2|\;\exists\; e_i\in E'_1,\;(e_i,e_j)\in \mathcal{C}_{\theta}\},
\end{aligned}
\end{equation}
where without loss of generality, we assume $|E''_1|<|E''_2|$. The alignment candidate set $\mathcal{C}_{\theta}$ inevitably involves some conflicting alignments. To warrant one-to-one alignment, we propose to model entity alignment as an OT process, i.e., transporting each entity $e_i \in E''_1$ to a unique entity $e_j \in E''_2$, with the minimal overall transport cost. 
Naturally, the rectified distance can be used to define the transport cost across KGs:
\begin{equation}
C(e_i,e_j)=
\tilde{d}(e_i,e_j),
\; e_i\in E''_1, e_j\in E''_2,
\end{equation}  
where $C(e_i,e_j)$ is the transport cost between entity $e_i$ and $e_j$.
The transport plan is in the form of a bijection $T: E''_1 \rightarrow E''_2$. In other words, each entity $e_{i} \in E''_1$ has exactly one transport to the target entity $T(e_{i}) \in E''_2$.  
The goal is to find the optimal transport plan $T^*$ that minimises the \dk{overall} transport cost: 
\begin{equation}
T^*=\mathop{\arg\min}\limits_{T} \sum_{e_i\in E''_1} C(e_i, T(e_i)).
\label{eq:general_ot_problem}
\end{equation}
The objective can be reformulated as: 
\begin{equation}
\begin{aligned}
&\mathop{\arg\min}\limits_{P\in\{0,1\}^{|E''_1|\times|E''_2|}} \langle P, C\rangle_F, \text{ subject to:} \\
&\sum_{e_j\in E''_2} P_{e_i,e_j} = 1,\;\sum_{e_i\in E''_1} P_{e_i,e_j} \leq 1, 
\label{eq:specific_ot_problem}
\end{aligned}
\end{equation}where $\langle \cdot,\cdot \rangle_F$ is the Frobenius inner product between two matrices. 
$P\in \{0,1\}^{|E''_1| \times |E''_2|}$ is the \dk{transport indicating} matrix and each element $P_{e_i,e_j}$ denotes \dk{whether} $e_i \in E''_1$ is aligned to $e_j \in E''_2$ \dk{with 1 for true and 0 for false}. 
To achieve one-to-one alignments \dk{across $E''_1$ and $E''_2$ in two KGs, with $|E''_1|<|E''_2|$}, the summation of each row in $P$ is constrained to 1, \dk{while} the summation of each column is bounded by 1. 

\dk{To solve the OT problem above, some exact algorithms have been proposed, such as the Branch and Bound algorithm~\cite{laporte1992traveling}
.}
The exact algorithms guarantee to find a globally optimal transport plan but with prohibitively high computational cost for iterative pseudo labeling. Hence, we propose to use a greedy algorithm~\cite{preis1999linear
} as an efficient yet accurate approximation to exact algorithms, which is proven to have an at least 1/2 approximation ratio as compared to exact algorithms~\cite{
nemhauser1978analysis}.


The overall process of our greedy algorithm for OT-based pseudo labeling is given in Algorithm~\ref{algorithm_greedy}. 
In Step 1, the pseudo-labeled alignment set $\hat{\mathbb{L}}_e$ and its increment $\Delta\hat{\mathbb{L}}_e$ are initialized as $\emptyset$.
The greedy algorithm first expands $\Delta\hat{\mathbb{L}}_e$  with the bounded shortest distance principle in Steps 2-5.
In Steps 6-10, the alignment conflicts in $\Delta\hat{\mathbb{L}}_e$ are eliminated through checking every two conflicting alignment pairs. 
As entity pairs can be sorted according to a lexicographic order, the operation can be finished in linear time.
Newly aligned entity pairs are then removed from $E'_1$ and $E'_2$ in Steps 11-14. In Step 15, $\hat{\mathbb{L}}_e$ is expanded with $\Delta\hat{\mathbb{L}}_e$ and $\Delta\hat{\mathbb{L}}_e$ is set to $\emptyset$. 
We then repeat the entity alignment augmentation process on the updated $E'_1$ and $E'_2$ until no updates in $\hat{\mathbb{L}}_e$ in Step 16. 
Finally, the greedy algorithm returns pseudo-labeled alignment pairs $\hat{\mathbb{L}}_e$.
Take the number of iterations in Step 16 as a constant, the overall time complexity of Algorithm 1 is $O(|E_1|\cdot |E_2|)$.


\subsection{Model Training for Entity Alignment}
After determining pseudo-labeled alignment pairs $\hat{\mathbb{L}}_e$ 
, the alignment seeds are augmented as: $\mathbb{L}_e \leftarrow \mathbb{L}^0_e \cup \hat{\mathbb{L}}_e$. 
Accordingly, we define the entity alignment loss as:
\begin{equation}
L = \sum_{(e_{i}, e_{j}) \in \mathbb{L}_e}\sum_{(e_{i'}, e_{j'}) \in \mathbb{L}^{\prime}_{e}} R(e_{i}, e_{j})\cdot 
[d(e_{i}, e_{j})-d(e_{i'}, e_{j'})+\gamma]_+,
\label{eq:soft_loss}
\end{equation}
where \dk{$[\cdot]_+$ denotes $\max(0,\cdot)$}, $\mathbb{L}_{e}^{\prime}$ is the set of sampled negative entity alignment pairs not included in $\mathbb{L}_e$, $\gamma$ is a positive margin hyper-parameter, and $d(\cdot,\cdot)$ is the embedding distance between two entities, as defined in Eq. (\ref{eq:dist}). $R(e_{i}, e_{j})\in (0,1]$ is the soft alignment score, i.e., the reliability score of each alignment pair $(e_{i}, e_{j}) \in \mathbb{L}_e$. For any prior aligned entity pair, $R(e_{i}, e_{j})=1$. For the augmented alignment,
\begin{equation}
R(e_{i}, e_{j}) = \sigma(w\cdot\theta - \tilde{d}(e_{i}, e_{j})),
\label{eq:reliability}
\end{equation}
where $\sigma(\cdot)$ is the sigmoid function, $\theta$ is the threshold \dk{used to determine alignment candidates}, and $w \in (0,1]$ is a hyper-parameter that controls the lower bound of $R$.

To obtain the negative entity alignment set $\mathbb{L}^{\prime}_{e}$, we adopt a \textit{adaptive negative sampling} strategy, i.e., for each positive entity pair $(e_i,e_j)$ in augmented alignment set $\mathbb{L}_e$, we select $K$ nearest entities of $e_i$ \dk{measured by} the embedding distance in Eq.(\ref{eq:dist}) to replace $e_j$ and form $K$ negative counterparts $(e_i,e_{j'})$. This strategy helps push entities in misaligned entity pairs far away from each other in the embedding space. 

Note that, as a special case when there are no prior alignment seeds, initialized entity embeddings without training are used for pseudo labeling instead.


\begin{algorithm}[t]
\caption{\dk{Optimal Transport based Pseudo Labeling with Greedy Algorithm}\label{algorithm_greedy}}
\KwData{Two unaligned entity sets $E'_1 \subseteq E_1$ and $E'_2 \subseteq E_2$. 
The rectified distance $\tilde{d}(\cdot,\cdot)$ and distance threshold $\theta$.}
\KwResult{Pseudo-labeled alignment pair \dk{set} $\hat{\mathbb{L}}_e$.}
Initialize $\hat{\mathbb{L}}_e \leftarrow\emptyset$ and $\Delta\hat{\mathbb{L}}_e \leftarrow\emptyset$\;
\For{each $e_i \in E'_1$}{
\dk{Find} $e_j \in E'_2$ with \dk{minimal} \dk{$\tilde{d}(e_i,e_j)$}\;
\If{$\tilde{d}(e_i,e_j)<\theta$}
{\dk{Expand} $\Delta\hat{\mathbb{L}}_e \leftarrow \Delta\hat{\mathbb{L}}_e \cup \{(e_i, e_j)\} $\;}
}
\dk{
\For{each $(e_i,e_j),(e_{i^{\prime}},e_j)\in\Delta\hat{\mathbb{L}}_e$}
{
\eIf{$\tilde{d}(e_i,e_j)\leq \tilde{d}(e_{i^{\prime}},e_j)$}
{Update $\Delta\hat{\mathbb{L}}_e \leftarrow \Delta\hat{\mathbb{L}}_e\setminus \{(e_{i^{\prime}},e_j)\}$\;}
{Update $\Delta\hat{\mathbb{L}}_e \leftarrow \Delta\hat{\mathbb{L}}_e\setminus \{(e_{i},e_j)\}$\;}
}
\For{each $(e_i,e_j)\in\Delta\hat{\mathbb{L}}_e$ with $e_i\in E'_{1}$}
{Update $E'_1 \leftarrow E'_1\setminus\{e_i\}$\;}
\For{each $(e_i,e_j)\in\Delta\hat{\mathbb{L}}_e$ with $e_j\in E'_{2}$}
{Update $E'_2 \leftarrow E'_2\setminus\{e_j\}$\;}}
Expand $\hat{\mathbb{L}}_e \leftarrow \hat{\mathbb{L}}_e \cup \Delta\hat{\mathbb{L}}_e$\ and set $\Delta\hat{\mathbb{L}}_e \leftarrow \emptyset$\;
Repeat Step\dk{s} 2-\dk{15} until no \dk{updates} in $\hat{\mathbb{L}}_e$\;
\Return{} Pseudo-labeled alignment pair \dk{set} $\hat{\mathbb{L}}_e$.
\end{algorithm}

With iterative pseudo labeling and model training, the final learned entity embeddings $\bm{h}_e$ are informative enough to measure the similarity between entities. We thus directly use the embedding distance defined in Eq.(\ref{eq:dist}) to infer aligned entities. Given two sets of unaligned entities, $E'_1\subseteq E_1$ and $E'_2\subseteq E_2$, for each entity $e_{i}\in E'_1$, we find the entity $e_{j}\in E'_2$ having the smallest embedding distance to $e_{i}$ as its alignment.

\section{Experiments}

\subsection{Datasets and Baselines}
We evaluate the performance of our CPL-OT\footnote{Source code: \url{https://github.com/qdin4048/CPL-OT}} method on two benchmark datasets, DBP15K~\cite{sun2017cross} and SRPRS~\cite{guo2019learning}. The statistics of both datasets are provided in Table~\ref{tab:dataset}.

\renewcommand\arraystretch{0.7}
\begin{table}[H]
  \begin{center}
   \caption{Statistics of datasets}
   \label{tab:dataset}
\begin{tabular}{c|c|c c c}
      \toprule 
	  \multicolumn{2}{c|}{\textbf{Datasets}} & \textbf{Entities} & \textbf{Relations} & \textbf{Rel.triplets} \\
      \midrule 
	  \multirow{2}{*}{DBP15K\textsubscript{ZH\_EN}} & Chinese & 66,469 & 2,830 & 153,929 \\
                                                                         & English & 98,125 & 2,317 & 237,674 \\
      \midrule 
	  \multirow{2}{*}{DBP15K\textsubscript{JA\_EN}} & Japanese & 65,744 & 2,043 & 164,373 \\
                                                                        & English & 95,680 & 2,096 & 233,319 \\
      \midrule 
	  \multirow{2}{*}{DBP15K\textsubscript{FR\_EN}} & French & 66,858 & 1,379 & 192,191\\
                                                                        & English & 105,889 & 2,209 & 278,590 \\
      \midrule 
	  \multirow{2}{*}{SRPRS\textsubscript{EN\_FR}} & English & 15,000 & 221 & 36,508\\
                                                                        & French & 15,000 & 177 & 33,532 \\
      \midrule 
	  \multirow{2}{*}{SRPRS\textsubscript{EN\_DE}} & English & 15,000 & 222 & 38,363\\
                                                                        & German &  15,000 & 120 & 37,377 \\
      \bottomrule 
    \end{tabular}
  \end{center}
\end{table}


\renewcommand{\arraystretch}{0.6}
\begin{table*}[h!]
\centering
\tabcolsep 3pt
   \caption{Performance comparison on DBP15K and SRPRS}
    \label{tab:comparison}
    \scalebox{0.9}{
\begin{tabular}{l|c c c|c c c|c c c|c c c|c c c}
      \toprule 
      \multirow{2}{*}{Models} & \multicolumn{3}{c|}{DBP15K\textsubscript{ZH\_EN}} & \multicolumn{3}{c|}{DBP15K\textsubscript{JA\_EN}} & \multicolumn{3}{c|}{DBP15K\textsubscript{FR\_EN}} & \multicolumn{3}{c|}{SRPRS\textsubscript{EN\_FR}} & \multicolumn{3}{c}{SRPRS\textsubscript{EN\_DE}}\\ 
      \cmidrule(l{0em}r{0em}){2-16} 
        &Hit@1 & Hit@10 & MRR & Hit@1  & Hit@10 & MRR & Hit@1 & Hit@10 & MRR & Hit@1 & Hit@10 & MRR & Hit@1 & Hit@10 & MRR\\
      \midrule 
       \multicolumn{16}{c}{30\% Prior Alignment Seeds}\\
      \midrule 
      MtransE~\cite{chen2016multilingual} & 20.9 & 51.2 & 0.31 & 25.0 & 57.2 & 0.36 & 24.7 & 57.7 & 0.36 & 21.3 & 44.7 & 0.29 & 10.7 & 24.8 & 0.16\\
      JAPE-Stru~\cite{sun2017cross}& 37.2 & 68.9 & 0.48 & 32.9 & 63.8 & 0.43 & 29.3 & 61.7 & 0.40 & 24.1 & 53.3 & 0.34 & 30.2 & 57.8 & 0.40\\ 
      GCN-Stru~\cite{wang2018cross} & 39.8 & 72.0 & 0.51 & 40.0 & 72.9 & 0.51 & 38.9 & 74.9 & 0.51 & 24.3 & 52.2 & 0.34 & 38.5 & 60.0 & 0.46\\ 
      \midrule 
      IPTransE~\cite{zhu2017iterative} & 33.2 & 64.5 & 0.43 & 29.0 & 59.5 & 0.39 & 24.5 & 56.8 & 0.35 & 12.4 & 30.1 & 0.18 & 13.5 & 31.6 & 0.20\\
      BootEA~\cite{sun2018bootstrapping} & 61.4 & 84.1 & 0.69 & 57.3 & 82.9 & 0.66 & 58.5 & 84.5 & 0.68 & 36.5 & 64.9 & 0.46 & 50.3 & 73.2 & 0.58\\
      MRAEA~\cite{mao2020mraea} & 75.7 & 93.0 & 0.83 & 75.8 & 93.4 & 0.83 & 78.0 & 94.8 & 0.85 & 46.0 & 76.8 & 0.56 & 59.4 & 81.5 & 0.66\\
      \midrule 
      GCN-Align~\cite{wang2018cross} & 43.4 & 76.2 & 0.55 & 42.7 & 76.2 & 0.54 & 41.1 & 77.2 & 0.53 & 29.6 & 59.2 & 0.40 & 42.8 & 66.2 & 0.51\\
      JAPE~\cite{sun2017cross} & 41.4 & 74.1 & 0.53 & 36.5 & 69.5 & 0.48 & 31.8 & 66.8 & 0.44 & 24.1 & 54.4 & 0.34 & 26.8 & 54.7 & 0.36\\
      HMAN~\cite{yang2019aligning} & 56.1 & 85.9 & 0.67 & 55.7 & 86.0 & 0.67 & 55.0 & 87.6 & 0.66 & 40.0 & 70.5 & 0.50 & 52.8 & 77.8 & 0.62\\      
      RDGCN~\cite{wu2019relation} & 69.7 & 84.2 & 0.75 & 76.3 & 89.7 & 0.81 & 87.3 & 95.0 & 0.90 & 67.2 & 76.7 & 0.71 & 77.9 & 88.6 & 0.82\\
      HGCN~\cite{wu2019jointly} & 70.8 & 84.0 & 0.76 & 75.8 & 88.9 & 0.81 & 88.8 & 95.9 & 0.91 & 67.0 & 77.0 & 0.71 & 76.3 & 86.3 & 0.80\\
      RNM~\cite{zhu2021relation2} & \underline{84.0} & \underline{91.9} & \underline{0.87} & \underline{87.2} & \underline{94.4} & \underline{0.90} & 93.8 & \underline{98.1} & \underline{0.95} & 92.5 & \underline{96.2} & \underline{0.94} & 94.4 & \underline{96.7} & \underline{0.95}\\
      CEA~\cite{Zeng0T020} & 78.7 & - & - & 86.3 & - & - & \underline{97.2} & - & - & \underline{96.2} & - & - & \underline{97.1} & - & - \\
     \midrule 
      CPL-OT & \textbf{92.7} & \textbf{96.4} & \textbf{0.94} & \textbf{95.6} & \textbf{98.3} & \textbf{0.97} & \textbf{99.0} & \textbf{99.4} & \textbf{0.99}& \textbf{97.4} & \textbf{98.8} & \textbf{0.98} & \textbf{97.4} & \textbf{98.9} & \textbf{0.98}\\ 
      \midrule 
      \multicolumn{16}{c}{No Prior Alignment Seeds}\\
      \midrule 
      MRAEA~\cite{mao2020mraea} & 77.8 & 83.2 & - & 88.9 & 92.7 & - & 95.0 & 97.0 & - & \underline{93.4} & \underline{96.0} & \underline{0.92} & \underline{94.9} & \underline{96.3} & \underline{0.92} \\
      SelfKG~\cite{liu2022selfkg} & \underline{82.9} & \underline{91.9}  & - & \underline{89.0} & \underline{95.3} & - & \underline{95.9} & \textbf{99.2} & - & - & - & - & - & - & -\\
      \midrule 
      CPL-OT & \textbf{91.1} & \textbf{95.0} & \textbf{0.93} & \textbf{94.4} & \textbf{97.7} & \textbf{0.96} & \textbf{98.6} & \underline{99.1} & \textbf{0.99} & \textbf{97.1} & \textbf{98.7} & \textbf{0.98} & \textbf{97.2} & \textbf{98.6} & \textbf{0.98}\\ 
      \bottomrule 
    \end{tabular}}
\end{table*}


For evaluation, we compare CPL-OT with 12 state-of-the-art entity alignment models categorized into three groups: 

\begin{itemize}[leftmargin=*]
\item Models that leverage KG structure only, including MTransE~\cite{chen2016multilingual}, 
JAPE~\cite{sun2017cross} and GCN-Align~\cite{wang2018cross} in their structure-only variants denoted as JAPE-Stru and GCN-Stru.


\item Models based on bootstrapping, including IPTransE~\cite{zhu2017iterative}, BootEA~\cite{sun2018bootstrapping}, and MRAEA~\cite{mao2020mraea}; 

\item Models that use auxiliary  information with KG structure, including GCN-Align~\cite{wang2018cross}, JAPE~\cite{sun2017cross}, RDGCN~\cite{wu2019relation}, HGCN~\cite{wu2019jointly}, RNM~\cite{zhu2021relation2}, HMAN~\cite{yang2019aligning}, CEA~\cite{zeng2020collective}, MRAEA~\cite{mao2020mraea} in its unsupervised variant, and SelfKG~\cite{liu2022selfkg} in its variant using translated version of word embeddings. 
\end{itemize}

We use Hit@$k$ ($k=1,10$) and Mean Reciprocal Rank (MRR) as evaluation metrics.
Higher Hit@$k$ and MRR scores indicate better entity alignment performance.

\subsection{Experimental Setup}
We follow the conventional 30\%-70\% training-test split on DBP15K and SRPRS. 
We use semantic meanings of entity names to construct entity features. On DBP15K with big linguistic barriers, we first use Google Translate to translate non-English entity names into English, then look up 768-dimensional word embeddings pre-trained by BERT~\cite{devlin2018bert}. On SRPRS with small linguistic barriers, we directly look up word embeddings without translation. For each entity, we aggregate TF-IDF-weighted word embeddings to form its feature vector.

CPL-OT uses the following parameter settings: $d=300$, $\lambda=10$, $w = 0.25$, $\theta=4$, $\gamma=1$ and $K=125$. 
For BERT pre-trained word embeddings, we use a PCA-based technique~\cite{raunak2019effective} to reduce feature dimension from 768 to 300 with minimal information loss. The batch size is set to 256 and the number of training epochs is set to 80. The Adam optimizer is used with a learning rate of 0.001 and 0.00025 on DBP15K and SRPRS, respectively. 
All experiments are run in Pytorch on an RTX 2080 Ti (11GB memory) GPU.


We re-produce the results of RNM and the unsupervised variant of MRAEA on SRPRS using their open-sourced code. Since entity features are not originally provided by SRPRS, we directly use our BERT-based entity features weighted by TF-IDF for re-implementation. The results of MRAEA on both benchmarks, RNM on DBP15K, and SelfKG on DBP15K are obtained from their original papers. Results of other baselines are obtained from~\cite{zhao2020experimental}. For the proposed CPL-OT, we repeat the experiment five times and report the average results.

\subsection{Performance Comparison with State-of-the-Art}
Table~\ref{tab:comparison} compares 
different models on five cross-lingual datasets from DBP15K and SRPRS. The results are reported under two settings: using 30\% prior alignment seeds, and with no prior alignment seeds, where all aligned pairs are used for testing.  
The best and second best performing methods are marked in \textbf{boldface} and \underline{underlined}, respectively. 


\subsubsection{30\% Prior Alignment Seeds}
In this setting, CPL-OT significantly beats all existing models on five datasets. In particular, on DBP15K\textsubscript{ZH\_EN}, CPL-OT outperforms the second best model by nearly 9\% in terms of Hit@1. We note that there are clear overall performance gaps among the five datasets, where the lowest accuracy is achieved on DBP15K\textsubscript{ZH\_EN} due to its large linguistic barriers. Thus, we regard entity alignment on DBP15K\textsubscript{ZH\_EN} as the most challenging task.

\subsubsection{No Prior Alignment Seeds}
In the case of no prior alignment seeds, 
CPL-OT also achieves superior results, significantly outperforming MRAEA and SelfKG. Benefiting from its conflict-aware pseudo-labelling, CPL-OT even outperforms all baselines using 30\% prior alignment seeds. When prior alignment seeds are reduced from 30\% to zero, the performance of CPL-OT retains stable
. The maximum drop of Hit@1 for CPL-OT is only 1.6\% on DBP15K\textsubscript{ZH\_EN}. 

\renewcommand{\arraystretch}{0.7}
\begin{table*}[h!]
\tabcolsep 1.7pt
  \begin{center}
   \caption{Ablation study of CPL-OT}
    \label{tab:ablation}
\begin{tabular}{l|c c c|c c c|c c c|c c c|c c c}
      \toprule 
      \multirow{2}{*}{Models} & \multicolumn{3}{c|}{DBP15K\textsubscript{ZH\_EN}} & \multicolumn{3}{c|}{DBP15K\textsubscript{JA\_EN}} & \multicolumn{3}{c|}{DBP15K\textsubscript{FR\_EN}} & \multicolumn{3}{c|}{SRPRS\textsubscript{EN\_FR}} & \multicolumn{3}{c}{SRPRS\textsubscript{EN\_DE}}\\ 
      \cmidrule(l{0em}r{0em}){2-16} 
        &Hit@1 & Hit@10 & MRR & Hit@1  & Hit@10 & MRR & Hit@1 & Hit@10 & MRR & Hit@1 & Hit@10 & MRR & Hit@1 & Hit@10 & MRR\\
      \midrule 
       \multicolumn{16}{c}{30\% Prior Alignment Seeds}\\
      \midrule 
      Full Model & 92.7 & \textbf{96.4} & \textbf{0.94} & \textbf{95.6} & \textbf{98.3} & \textbf{0.97} & 99.0 & 99.4 & \textbf{0.99} & 97.4 & \textbf{98.8} & \textbf{0.98} & \textbf{97.4} & \textbf{98.9} & \textbf{0.98}\\ 
      w.o. Global-level Rel. Aggr. & 89.3 & 94.1 & 0.91 & 94.2 & 97.2 & 0.95 & \textbf{99.1} & \textbf{99.6} & 0.99 & 96.3 & 97.6 & 0.97 & 96.7 & 98.0 & 0.97\\
      w.o. Emb. Dist. Rect. & 84.2 & 91.9 & 0.87 & 90.4 & 95.6 & 0.92 & 96.9 & 98.4 & 0.98 & 95.0 & 97.3 & 0.96 & 96.4 & 98.2 & 0.97\\ 
      w.o. Conflict-aware OT & 91.7 & 95.3 & 0.93 & 94.8 & 97.8 & 0.96 & 98.5 & 99.2 & 0.99 & 96.9 & 98.5 & 0.98 & 96.8  & 98.6 & 0.98 \\
      w.o. Soft Align. & \textbf{92.9} & 96.2 & 0.94 & 95.2 & 98.0 & 0.96 & 98.9 & 99.4 & 0.99 & \textbf{97.6} & 98.7 & 0.98 & 97.4 & 98.7 & 0.98\\ 
      \midrule 
      \multicolumn{16}{c}{No Prior Alignment Seeds}\\
      \midrule 
      Full Model & \textbf{91.1} & \textbf{95.0} & \textbf{0.93} & \textbf{94.5} & \textbf{97.6} & \textbf{0.96} & 98.6 & \textbf{99.2} & \textbf{0.99} & \textbf{97.1} & 98.6 & \textbf{0.98} & \textbf{97.2} & 98.4 & \textbf{0.98}\\ 
      w.o. Global-level Rel. Aggr. & 88.8 & 93.0 & 0.90 & 93.4 & 96.9 & 0.95 & 98.6 & 99.5 & 0.99 & 96.2 & 97.5 & 0.97 & 96.1 & 97.5 & 0.97\\
      w.o. Emb. Dist. Rect.& 70.3 & 77.7 & 0.73 & 80.9 & 87.6 & 0.83 & 94.6 & 96.2 & 0.95 & 91.5 & 93.6 & 0.92 & 93.6 & 96.1 & 0.95\\ 
      w.o. Conflict-aware OT & 90.0 & 94.0 & 0.91 & 93.4 & 96.7 & 0.95 & 98.1 & 98.9 & 0.98 & 96.5 & 98.2 & 0.97 & 96.5 & 98.3 & 0.97\\      
      w.o. Soft Align. & 90.7 & 94.7 & 0.92 & 94.0 & 97.4 & 0.95 & \textbf{98.7} & 99.3 & 0.99 & 96.8 & \textbf{98.7} & 0.98 & 97.0 & \textbf{98.7} & 0.98\\ 
      \bottomrule 
    \end{tabular}
  \end{center}

\end{table*}

\subsection{Ablation Study}
We conduct a series of ablation study to investigate the importance of different components of the proposed CPL-OT model on both settings of 30\% prior alignment seeds and no prior alignment seeds. Table~\ref{tab:ablation} compares the full CPL-OT model with its ablated variants, with the best performance highlighted by \textbf{boldface}. From Table~\ref{tab:ablation}, we can find the full CPL-OT model overall performs the best in all cases. 

\subsubsection{Ablation on Global-Level Relation Aggregation} 
Without global-level relation aggregation (w.o. Global-level Rel. Aggr.), entities tend to be over-smoothed by neighboring entity features, thereby incurring more conflicts during pseudo labeling and degrading model performance on both settings.

\subsubsection{Ablation on Embedding Distance Rectification} 
As relational neighborhood matching can well complement embedding distance for entity alignment, by providing additional evidence contributed by aligned neighboring entities and relations. Ablating this component (w.o. Emb. Dist. Rect.) leads to a dramatic performance drop on both settings.

\subsubsection{Ablation on Conflict-aware Alignment with OT} 
After replacing OT-based alignment with a naive alignment strategy that simply uses Eq.(\ref{eq:naive_align}) to preserve only the most convincing aligned entity pairs for handling conflicts (w.o. Conflict-aware OT)
, the model fails to pseudo-label sufficient correct alignments, resulting in inferior performance on both settings.


\subsubsection{Ablation on Soft Alignment} 
On the setting with 30\% prior alignment seeds, 
ablating soft alignment (w.o. Soft Align.) has comparable performance to the full model. However, on the setting with no prior alignment seeds, pseudo labeling is prone to errors 
due to the lack of high-quality entity embeddings, so the ablation of soft alignment degrades model performance on DBP15K\textsubscript{ZH\_EN} and DBP15K\textsubscript{JA\_EN}.

\section{Conclusion}
We proposed a novel conflict-aware pseudo labeling model (CPL-OT) for entity alignment across KGs. 
CPL-OT augments the training data with sufficiently reliable alignment pairs via an OT modeling for alleviating alignment conflicts. Competitive performance of CPL-OT on two benchmark datasets demonstrates the superiority of OT-based pseudo-labeling strategy and its great potential for entity alignment in KGs.

\bibliographystyle{IEEEtran}
\bibliography{reference}

\begin{thebibliography}{10}
\providecommand{\url}[1]{#1}
\csname url@samestyle\endcsname
\providecommand{\newblock}{\relax}
\providecommand{\bibinfo}[2]{#2}
\providecommand{\BIBentrySTDinterwordspacing}{\spaceskip=0pt\relax}
\providecommand{\BIBentryALTinterwordstretchfactor}{4}
\providecommand{\BIBentryALTinterwordspacing}{\spaceskip=\fontdimen2\font plus
\BIBentryALTinterwordstretchfactor\fontdimen3\font minus
  \fontdimen4\font\relax}
\providecommand{\BIBforeignlanguage}[2]{{%
\expandafter\ifx\csname l@#1\endcsname\relax
\typeout{** WARNING: IEEEtran.bst: No hyphenation pattern has been}%
\typeout{** loaded for the language `#1'. Using the pattern for}%
\typeout{** the default language instead.}%
\else
\language=\csname l@#1\endcsname
\fi
#2}}
\providecommand{\BIBdecl}{\relax}
\BIBdecl

\bibitem{wang2018cross}
Z.~Wang, Q.~Lv, X.~Lan, and Y.~Zhang, ``Cross-lingual knowledge graph alignment
  via graph convolutional networks,'' in \emph{EMNLP}, 2018, pp. 349--357.

\bibitem{kipf2016semi}
T.~N. Kipf and M.~Welling, ``Semi-supervised classification with graph
  convolutional networks,'' in \emph{ICLR}, 2017.

\bibitem{wu2019jointly}
Y.~Wu, X.~Liu, Y.~Feng, Z.~Wang, and D.~Zhao, ``Jointly learning entity and
  relation representations for entity alignment,'' in \emph{EMNLP/IJCNLP},
  2019, pp. 240--249.

\bibitem{wu2019relation}
Y.~Wu, X.~Liu, Y.~Feng, Z.~Wang, R.~Yan, and D.~Zhao, ``Relation-aware entity
  alignment for heterogeneous knowledge graphs,'' in \emph{IJCAI}, 2019, pp.
  5278--5284.

\bibitem{zhu2021relation2}
Y.~Zhu, H.~Liu, Z.~Wu, and Y.~Du, ``Relation-aware neighborhood matching model
  for entity alignment,'' in \emph{AAAI}, 2021, pp. 4749--4756.

\bibitem{srivastava2015highway}
R.~K. Srivastava, K.~Greff, and J.~Schmidhuber, ``Highway networks,''
  \emph{arXiv:1505.00387}, 2015.

\bibitem{sun2018bootstrapping}
Z.~Sun, W.~Hu, Q.~Zhang, and Y.~Qu, ``Bootstrapping entity alignment with
  knowledge graph embedding.'' in \emph{IJCAI}, 2018, pp. 4396--4402.

\bibitem{zhu2017iterative}
H.~Zhu, R.~Xie, Z.~Liu, and M.~Sun, ``Iterative entity alignment via knowledge
  embeddings,'' in \emph{IJCAI}, 2017, pp. 4258--4264.

\bibitem{mao2020mraea}
X.~Mao, W.~Wang, H.~Xu, M.~Lan, and Y.~Wu, ``Mraea: an efficient and robust
  entity alignment approach for cross-lingual knowledge graph,'' in
  \emph{WSDM}, 2020, pp. 420--428.

\bibitem{villani2009optimal}
C.~Villani, \emph{Optimal transport: old and new}.\hskip 1em plus 0.5em minus
  0.4em\relax Springer, 2009, vol. 338.

\bibitem{pei2019improving}
S.~Pei, L.~Yu, and X.~Zhang, ``Improving cross-lingual entity alignment via
  optimal transport.''\hskip 1em plus 0.5em minus 0.4em\relax IJCAI, 2019, pp.
  3231--3237.

\bibitem{chen2020graph}
L.~Chen, Z.~Gan, Y.~Cheng, L.~Li, L.~Carin, and J.~Liu, ``Graph optimal
  transport for cross-domain alignment,'' in \emph{ICML}, 2020, pp. 1542--1553.

\bibitem{xu2019gromov}
H.~Xu, D.~Luo, H.~Zha, and L.~C. Duke, ``Gromov-wasserstein learning for graph
  matching and node embedding,'' in \emph{ICML}, 2019, pp. 6932--6941.

\bibitem{laporte1992traveling}
G.~Laporte, ``The traveling salesman problem: An overview of exact and
  approximate algorithms,'' \emph{European Journal of Operational Research},
  vol.~59, no.~2, pp. 231--247, 1992.

\bibitem{preis1999linear}
R.~Preis, ``Linear time 1/2-approximation algorithm for maximum weighted
  matching in general graphs,'' in \emph{Annual Symposium on Theoretical
  Aspects of Computer Science}, 1999, pp. 259--269.

\bibitem{nemhauser1978analysis}
G.~L. Nemhauser, L.~A. Wolsey, and M.~L. Fisher, ``An analysis of
  approximations for maximizing submodular set functions—i,''
  \emph{Mathematical Programming}, vol.~14, no.~1, pp. 265--294, 1978.

\bibitem{sun2017cross}
Z.~Sun, W.~Hu, and C.~Li, ``Cross-lingual entity alignment via joint
  attribute-preserving embedding,'' in \emph{ISWC}, 2017, pp. 628--644.

\bibitem{guo2019learning}
L.~Guo, Z.~Sun, and W.~Hu, ``Learning to exploit long-term relational
  dependencies in knowledge graphs,'' in \emph{ICML}, 2019, pp. 2505--2514.

\bibitem{chen2016multilingual}
M.~Chen, Y.~Tian, M.~Yang, and C.~Zaniolo, ``Multilingual knowledge graph
  embeddings for cross-lingual knowledge alignment,'' in \emph{IJCAI}, 2017,
  pp. 1511--1517.

\bibitem{yang2019aligning}
H.-W. Yang, Y.~Zou, P.~Shi, W.~Lu, J.~Lin, and X.~Sun, ``Aligning cross-lingual
  entities with multi-aspect information,'' in \emph{EMNLP-IJCNLP}, 2019, pp.
  4431--4441.

\bibitem{Zeng0T020}
W.~Zeng, X.~Zhao, J.~Tang, and X.~Lin, ``Collective entity alignment via
  adaptive features,'' in \emph{ICDE}, 2020, pp. 1870--1873.

\bibitem{liu2022selfkg}
X.~Liu, H.~Hong, X.~Wang, Z.~Chen, E.~Kharlamov, Y.~Dong, and J.~Tang,
  ``Selfkg: Self-supervised entity alignment in knowledge graphs,'' in
  \emph{WWW}, 2022, pp. 860--870.

\bibitem{zeng2020collective}
W.~Zeng, X.~Zhao, J.~Tang, and X.~Lin, ``Collective entity alignment via
  adaptive features,'' in \emph{ICDE}, 2020, pp. 1870--1873.

\bibitem{devlin2018bert}
J.~Devlin, M.-W. Chang, K.~Lee, and K.~Toutanova, ``Bert: Pre-training of deep
  bidirectional transformers for language understanding,'' \emph{arXiv preprint
  arXiv:1810.04805}, 2018.

\bibitem{raunak2019effective}
V.~Raunak, V.~Gupta, and F.~Metze, ``Effective dimensionality reduction for
  word embeddings,'' in \emph{The 4th Workshop on Representation Learning for
  NLP}, 2019, pp. 235--243.

\bibitem{zhao2020experimental}
X.~Zhao, W.~Zeng, J.~Tang, W.~Wang, and F.~Suchanek, ``An experimental study of
  state-of-the-art entity alignment approaches,'' \emph{IEEE TKDE}, vol.~34,
  no.~6, pp. 2610--2625, 2020.

\end{thebibliography}

\end{document}